\documentclass[letterpaper, 10 pt, conference]{ieeeconf}  % Comment this line out if you need a4paper

\IEEEoverridecommandlockouts                              % This command is only needed if 
\usepackage{graphicx}
\usepackage{amsmath,amssymb}
\usepackage{algorithm}
\usepackage{algorithmicx}
\usepackage{algpseudocode}
\usepackage{url}
\usepackage[skip=2.5pt]{subcaption}
\usepackage{enumerate}

\usepackage{booktabs}
\usepackage{multirow}
\usepackage{makecell}
\usepackage{bbm}

\title{\LARGE \bf
Learning Bipedal Locomotion on Gear-Driven Humanoid Robot Using Foot-Mounted IMUs
}

\author{Sotaro Katayama$^{1}$, Yuta Koda$^{2}$, Norio Nagatsuka$^{2}$, and Masaya Kinoshita$^{1}$% <-this % stops a space
\thanks{$^{1}$Sony Group Corporation, Minato-ku, Tokyo, Japan, 108-0075
        {\tt\small sotaro.katayama@sony.com}}%
\thanks{$^{2}$Sony Interactive Entertainment Inc., Sony City 1-7-1, Konan, Minato-ku, Tokyo, 108-0075 Japan}
}

\begin{document}

\maketitle
\thispagestyle{empty}
\pagestyle{empty}

%%%%%%%%%%%%%%%%%%%%%%%%%%%%%%%%%%%%%%%%%%%%%%%%%%%%%%%%%%%%%%%%%%%%%%%%%%%%%%%%
\begin{abstract}
Sim-to-real reinforcement learning (RL) for humanoid robots with high-gear ratio actuators remains challenging due to complex actuator dynamics and the absence of torque sensors.
To address this, we propose a novel RL framework leveraging foot-mounted inertial measurement units (IMUs).
Instead of pursuing detailed actuator modeling and system identification, we utilize foot-mounted IMU measurements to enhance rapid stabilization capabilities over challenging terrains.
Additionally, we propose symmetric data augmentation dedicated to the proposed observation space and random network distillation to enhance bipedal locomotion learning over rough terrain.
We validate our approach through hardware experiments on a miniature-sized humanoid EVAL-03 over a variety of environments.
The experimental results demonstrate that our method improves rapid stabilization capabilities over non-rigid surfaces and sudden environmental transitions.
\end{abstract}

%%%%%%%%%%%%%%%%%%%%%%%%%%%%%%%%%%%%%%%%%%%%%%%%%%%%%%%%%%%%%%%%%%%%%%%%%%%%%%%%
\section{INTRODUCTION}
\begin{figure}[t]
  \centering
  \begin{minipage}{0.44\linewidth}
    \centering
    \includegraphics[scale=0.155]{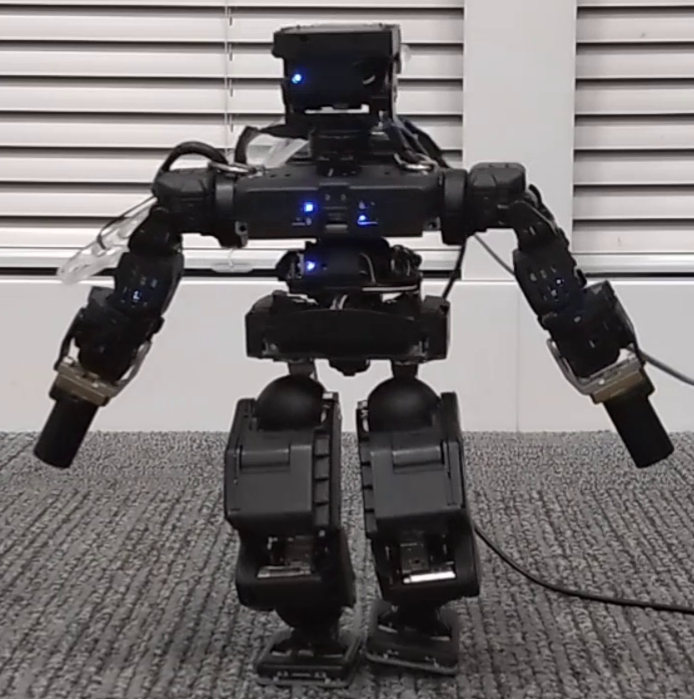}
  \end{minipage}
  \hspace{0.05\linewidth}
  \begin{minipage}{0.48\linewidth}
    \centering
    \includegraphics[scale=0.2]{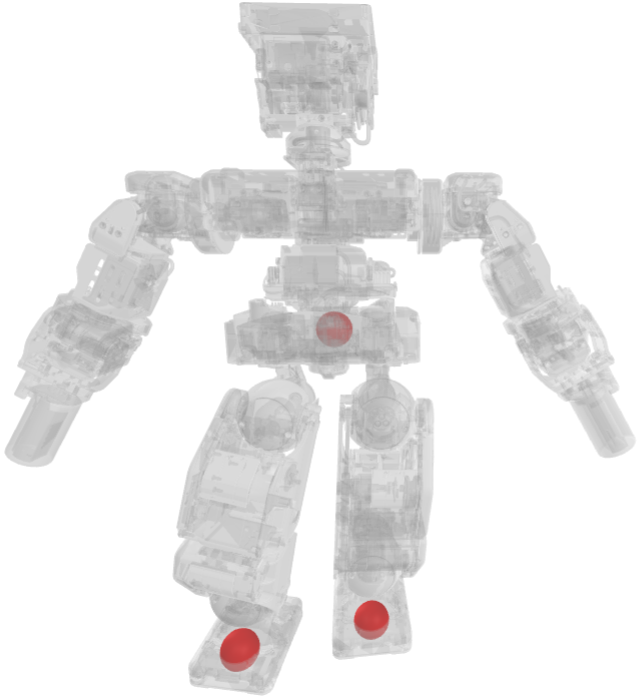}
  \end{minipage}
  \centering
  \includegraphics[width=0.9\columnwidth]{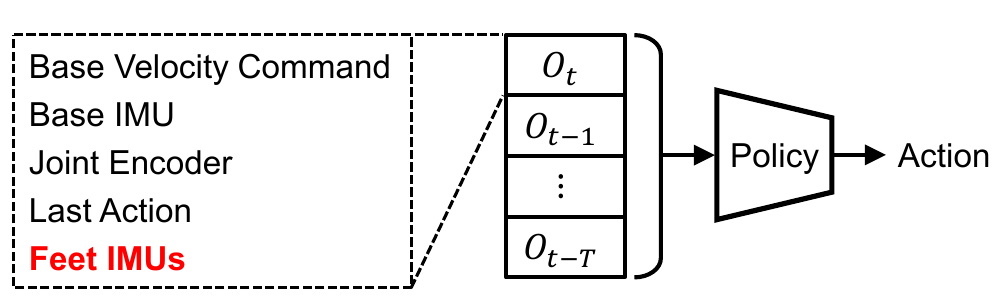}
  \caption{Upper: A photo (left) and kinematic model (right) of the gear-driven, miniature-sized humanoid robot EVAL-03. Lower: An overview of the proposed method. In the kinematic model, three IMUs mounted on the body, left foot, and right foot are illustrated as red spheres.}
  \label{fig:eval}
\end{figure}

Bipedal and humanoid robots have fascinated people for decades.
One of their anticipated roles is to replace human workers.
Humanoid robots, which have morphologies similar to humans, are expected to navigate environments accessible to humans and perform tasks that humans can accomplish.
Another significant application is in entertainment.
A pioneering example of entertainment robotics is the AIBO series \cite{aibo2024web}, a dog-like quadrupedal robot developed to interact with people.
QRIO \cite{fujita2024stories} is a small-sized humanoid robot that followed the same approach as AIBO but with biepdal locomotion.
BD-X \cite{grandia2024design}, a bipedal robot with a unique, character-like design, has demonstrated its entertainment applications in the real world.
EVAL-03, depicted in Fig. \ref{fig:eval}, was developed by Sony Interactive Entertainment to further explore the potential of robotics in entertainment.
In this work, we focus on enhancing the locomotion capabilities of EVAL-03, which has been limited to upper body movements, static posing, and walking on a flat plane without disturbances \cite{taylor2021learning}.

Reinforcement learning (RL) has demonstrated robust, dynamic, and natural locomotion capabilities \cite{li2024reinforcement, zhuang2024humanoid, haarnoja2024learning, unitree2025}.  
A key enabler of these advancements is zero-shot sim-to-real transfer, where training is conducted entirely in massively parallelized physics simulation frameworks \cite{makoviychuk2021isaac, rudin2022learning, mittal2023orbit}, and the learned policies are deployed directly on real hardware without fine-tuning.  
However, the success of this approach largely depends on mitigating the sim-to-real gap.  
One effective strategy for reducing this gap involves embedding an actuator network \cite{hwangbo2019learning} within the simulation.
This network, trained on real-world data, infers joint torques based on historical joint measurements in the simulation.
However, this method requires actuators equipped with torque sensors, which are often prohibitively expensive.  
An alternative approach is the use of direct or quasi-direct drive actuators \cite{katz2019mini, liao2024berkeley}, which enable accurate modeling of PD-controlled actuators within simulations.

Nonetheless, challenges persist in applying RL to robots with high-gear ratio actuators that lack torque sensors.
This is particularly relevant for low-cost, miniature-sized humanoid robots.
Such robots typically require high-gear ratios to amplify the capabilities of small-sized, low-power motors, often at the expense of increased backlash and joint friction, which leads to nonlinear torque-current relationships.
Moreover, these low-cost actuators generally do not have torque sensors, rendering actuator networks unavailable.
The ROBOTIS-OP3 \cite{robotis2024op3} exemplifies such a low-cost, miniature-sized humanoid robot and has been utilized in RL studies \cite{haarnoja2024learning, masuda2023sim}.
In \cite{haarnoja2024learning}, to mitigate the sim-to-real gap, the high-gain position control mode is employed while subsequently actuator parameters are identified.
Although the authors demonstrated agile soccer motions of the robot, these movements were limited to flat ground.  For effective locomotion over varied terrain, compliant joint control, i.e., low-gain joint position control, can be a crucial factor, as it aids in estimating contact states and terrain features, as reported in \cite{lee2020learning}.
To enable the ROBOTIS-OP3 to operate with compliant joint control, \cite{masuda2023sim} employed a more detailed actuator model identification.
Unfortunately, the results only demonstrated slow walking over a tilted plane, with no success on rough terrain or steps, despite the necessity for careful real data collection.
In a similar context, but for human-sized humanoid robots, \cite{xie2023learning} incorporates current feedback to account for torque-tracking errors of the actuators.
However, this approach still necessitates the identification of accurate motor parameters, such as motor armature and friction characteristics.
Upon these studies, we pose the question: can we improve sim-to-real transfer by introducing additional sensor observations?

In this paper, we propose the use of foot-mounted inertial measurement units (IMUs) for learning bipedal locomotion on a gear-driven humanoid robot.
As illustrated in Fig. \ref{fig:eval}, we utilize sensor measurements (linear accelerations and angular velocities) from foot-mounted IMUs as well as the base-mounted IMU within the blind locomotion learning framework \cite{lee2020learning}.
Additionally, we introduce symmetric data augmentation \cite{mayank2024symmetry} dedicated to the proposed observation space and random network distillation \cite{burda2019exploration, schwarke2023curiosity} to enhance the learning of bipedal locomotion over rough terrains.
We conducted hardware experiments on the gear-driven, miniature-sized humanoid robot EVAL-03 over a variety of environments including non-rigid surfaces and sudden environmental transitions.
Through the hardware experiments, we demonstrated that the proposed method improves rapid stabilization capabilities by leveraging the feet states measured by foot-mounted IMUs instead of employing detailed and careful system identification of the actuators as in \cite{masuda2023sim}.

\section{RELATED WORKS}\label{sec:relatedWorks}
\subsection{Sim-to-Real Transfer}\label{subsec:reviewsim2real}
Massively parallelized physical simulation frameworks \cite{makoviychuk2021isaac, rudin2022learning, mittal2023orbit} have enabled efficient collection of large amounts of training data.
However, when deploying policies trained solely in simulation to the real world, the sim-to-real gap—the discrepancy between the simulation model and real environment—can significantly affect the policy's performance.
A common approach to mitigate this gap is to employ domain randomization \cite{tobin2017domain} for parameters such as inertial properties (e.g., masses and centers of mass) and actuator characteristics (e.g., PD gains and friction).
For robots equipped with direct or quasi-direct drive actuators \cite{katz2019mini, liao2024berkeley}, simple domain randomization has proven effective, as these actuators can be accurately modeled using basic PD controllers in physical simulators.
However, when the actuator model deviates significantly from simple PD control—for instance, with substantial nonlinearity in the current-torque relationship—more detailed parameter identification becomes necessary \cite{tan2018sim, yu2019sim, masuda2023sim}.
This is particularly relevant for high-gear ratio actuators that exhibit backlash, joint friction, which leads to nonlinear torque-current relationships.
An alternative approach involves training neural networks to imitate real-world data \cite{hwangbo2019learning}.
However, this method is only viable when actuators are equipped with torque sensors, which are often prohibitively expensive.
Moreover, both detailed parameter identification and actuator networks require careful collection of real-world data to ensure sufficient coverage of possible observations.

\subsection{Leveraging Foot-Mounted IMUs}\label{subsec:reviewfeetIMU}
Foot-mounted IMUs have been utilized in human motion analysis \cite{perez2019real}.
However, in legged robotics, their application has been limited to a few studies \cite{xavier2023multi, yang2023multi}.
In \cite{xavier2023multi} and \cite{yang2023multi}, foot-mounted IMUs are employed to enhance state estimation in humanoid and quadrupedal robots, respectively.
The demonstrated effectiveness of foot-mounted IMU measurements in state estimation has inspired us to leverage them in RL-based locomotion control, as (partially observable) RL can encompass state estimation \cite{lee2020learning, kumar2021rma, miki2022learning}.

\section{METHOD}\label{sec:method}

\subsection{Reinforcement Learning of Bipedal Locomotion with Foot-Mounted IMUs}\label{subsec:rl}

Our method is based on Legged Gym \cite{rudin2022learning}, a model-free RL framework leveraging massively parallelized physical simulation \cite{makoviychuk2021isaac}.
The policy is conditioned on velocity commands comprising longitudinal, lateral, and yaw velocities ($v_{x, {\rm cmd}}$, $v_{y, {\rm cmd}}$, and $w_{z, {\rm cmd}}$, respectively).
As provided by Legged Gym, the policy is trained across various terrains, including slopes, rough surfaces, upward stairs, downward stairs, and discrete steps.
Each of these terrains is generated with 10 different difficulty levels, and we employ a constant curriculum similar to \cite{hwangbo2019learning}.

\begin{table}[t]
\centering
\caption{List of observation terms}
\begin{tabular}{lccc}
\hline
Input   & Obs. & Privileged obs. & Dim. \\
\hline
Base velocity command              & $\checkmark$  &  $\checkmark$  & 3    \\
Base IMU linear acceleration       & $\checkmark$  &  $\checkmark$  & 3    \\
Base IMU angular velocity          & $\checkmark$  &  $\checkmark$  & 3    \\
Base projected gravity             & $\checkmark$  &  $\checkmark$  & 3    \\
Joint positions                    & $\checkmark$  &  $\checkmark$  & 12   \\
Joint velocities                   & $\checkmark$  &  $\checkmark$  & 12   \\
Last actions                       & $\checkmark$  &  $\checkmark$  & 12   \\
{\bf Feet IMU accelerations}       & $\checkmark$  &  $\checkmark$  & 6    \\
{\bf Feet IMU angular velocities}  & $\checkmark$  &  $\checkmark$  & 6    \\
\hline
Noiseless joint positions          &   &  $\checkmark$  & 12    \\
Noiseless joint velocities         &   &  $\checkmark$  & 12    \\
Base linear velocity               &   &  $\checkmark$  & 3     \\
Noiseless base projected gravity   &   &  $\checkmark$  & 3     \\
Base push force                    &   &  $\checkmark$  & 3     \\
Base push torque                   &   &  $\checkmark$  & 3     \\
Feet contact forces                &   &  $\checkmark$  & 6     \\
Feet contact flags                 &   &  $\checkmark$  & 2     \\
Target feet contact flags          &   &  $\checkmark$  & 2     \\
Added base mass                    &   &  $\checkmark$  & 1     \\
COM displacement                   &   &  $\checkmark$  & 3     \\
Friction coefficient               &   &  $\checkmark$  & 1     \\
Restituition coefficient           &   &  $\checkmark$  & 1     \\
Height scan                        &   &  $\checkmark$  & 117   \\
\hline
\end{tabular}
\label{table:observations}
\end{table}

\begin{table}[t]
\centering
\caption{List of observation noise scales}
\begin{tabular}{lc}
\hline
Input & Noise scale \\
\hline
Base velocity command   & --    \\
Base/feet IMU acceleration        & 4.0   \\
Base/feet IMU angular velocity    & 0.1   \\
Base projected gravity  & 0.05  \\
Joint positions         & 0.05  \\
Joint velocities        & 1.0   \\
Last actions            & --    \\
\hline
\end{tabular}
\label{table:noise}
\end{table}

The observation space of the proposed method is detailed in Table \ref{table:observations}, and the observation noise for sim-to-real transfer is listed in Table \ref{table:noise}.
Notably, our observations include linear accelerations and angular velocities from IMUs mounted on the left and right feet.
Additionally, we incorporate the linear acceleration of the base IMU, which is absent in some existing studies on RL for locomotion.
We hypothesize that the foot-mounted IMUs enable direct and rapid measurement of feet states, which can improve capabilities of motion over a variety of terrains.

The action space consists of target joint positions for the low-level PD controller.
Since we often employ high-gain PD control for small-sized and low-cost actuators such as those in EVAL-03, the target joint position command must be smooth to avoid hardware damage.
To achieve this, following \cite{haarnoja2024learning}, we employ a low-pass filter:
\begin{equation}\label{eq:LPF}
  q_{\rm J, cmd} (t) = 0.8 \; q_{\rm J, cmd} (t - 1) + 0.2 \; a (t),
\end{equation}
where \( q_{\rm J, cmd} (t) \) is the target joint position at time \( t \) and \( a (t) \) is the latest scaled action at time \( t \).
The PD controller with the low pass filter update (\ref{eq:LPF}) run at 1000 Hz and and policy inference operate at 100 Hz in both simulation and real-robot scenarios.

\begin{table}[t]
\centering
\caption{List of reward function terms}
\begin{tabular}{lcc}
\hline
Reward Term & Expression & Weight \\
\hline
\hline
Lin. vel. tracking   & $\exp (- 1000 * (v_{xy} - v_{xy, {\rm cmd}})^2)$                         & 1.5 \\
Ang. vel. tracking   & $\exp (- 50 * (w_{\rm z} - w_{\rm z, {\rm cmd}})^2)$                     & 1.0 \\
\hline
Base rotation        & $g_{x, y} ^2$                                                            & -5.0 \\
Base height          & $(\min(h_{z} - h_{\rm target}, 0)) ^2$                                   & -0.2 \\
Lin. vel. penalty    & $v_{z} ^2$                                                               & -0.1 \\
Ang. vel. penalty    & $w_{xy} ^2$                                                              & -0.2 \\
\hline
Contact state        & $\mathbbm{1} ((F_z > 0.1) == {\rm TargetState})$ \cite{gu2024humanoid}   & 0.3 \\
Feet air time        & $T_{\rm air} - 0.5$ \cite{rudin2022learning}                             & 1.0 \\
Feet clearance       & $\mathbbm{1} (z_{\rm min} < z_{\rm swing} < z_{\rm max})$ \cite{gu2024humanoid} & 0.2 \\
Stance feet slip     & $v_{x, y} ^2 + w_z ^2$                                                 & -0.1 \\
\hline
Feet distance        & $\exp(\min(d_{\rm feet} - 0.05, 0))$                                     & -2.0 \\
Knee distance        & $\exp(\min(d_{\rm knee} - 0.05, 0))$                                     & -2.0 \\
Foot-knee distance   & $\exp(\min(d_{\rm foot-knee} - 0.05, 0))$                                     & -2.0 \\
\hline
Joint positions      & $|q_{J} - q_{\rm J, default}|$                                           & -0.1 \\
Joint velocities     & $\dot{q}_J ^2$                                                           & -5.0e-4  \\
Joint accelerations  & $\ddot{q}_J ^2$                                                          & -1.0e-7  \\
Joint torques        & $\tau_J ^2$                                                              & -5.0e-5  \\
Action rate          & $(a_{t} - a_{t -1}) ^2$                                                  & -0.01 \\
Action smoothness    & $(a_{t} - 2 a_{t -1} - a_{t - 2}) ^2$                                    & -0.01 \\
Termination          & $\mathbbm{1}_{\rm termination}$                                          & -200 \\
\hline
\end{tabular}
\label{table:reward}
\end{table}

In defining the reward function, we primarily follow the default reward structure provided by Legged Gym \cite{rudin2022learning}, while tuning the hyperparameters, particularly those regarding the robot's size.
We also introduce dedicated rewards for bipedal locomotion introduced in \cite{gu2024humanoid}: gait reward, swing-foot clearance reward, and penalties for self-collisions between left and right feet, while also adapting hyperparameters to the miniature-sized humanoid robot.
Our reward function terms are summarized in Table \ref{table:reward}.

\subsection{Sim-to-Real Considerations}\label{subsec:sim_to_real}

\begin{table}[t]
\centering
\caption{List of domain randomizations}
\begin{tabular}{lccr}
\hline
Parameter & Unit & Range & Operator \\
\hline
Joint position encoder offset  & rad            &  [-0.01, 0.01]   & Additive  \\
IMU accerarometer offset       & m/${\rm s^2}$  &  [-0.1, 0.1]     & Additive  \\
IMU gyroscope sensor offset    & rad/s          &  [-0.005, 0.005] & Additive  \\
Added base mass                & kg             &  [0.0, 0.2]      & Additive  \\
COM displacement               & m              &  [-0.05, 0.05]   & Additive  \\
Friction coefficient           & --             &  [0.1, 1.0]      & Scaling   \\
Restituition coefficient       & --             &  [0.0, 0.1]      & Additive  \\
$K_p$ factor                   & --             &  [0.9, 1.1]      & Scaling   \\
$K_d$ factor                   & --             &  [0.5, 1.5]      & Scaling   \\
System delay                   & ms             &  [0, 10]         & --        \\
\hline
\end{tabular}
\label{table:domainRand}
\end{table}

To enhance sim-to-real transfer, we employ domain randomizations such as additional base mass, center of mass (COM) displacements, and PD gains.
The parameters of the domain randomizations are listed in Table \ref{table:domainRand}.

It should be noted that, because the IMUs equipped on EVAL-03 are not high-grade, their accelerometer range is limited.
To close the sim-to-real gap, we replicate this limited sensor range in the simulation by clipping the linear acceleration observation terms.

Additionally, due to the low-cost motors, the actuators do not have an interface to provide joint velocities.
Therefore, in the hardware, we estimate joint velocities using finite differences of joint positions:
\begin{equation}\label{eq:jointVelocity}
\dot{q}_J (t) \simeq (q_J (t) - q_J (t - 1)) / \Delta t,
\end{equation}
at each 1000 Hz control loop, that is, $\Delta t = 1$ ms.
To close the sim-to-real gap, we also replicate this joint velocity estimation (\ref{eq:jointVelocity}) in simulation as the joint velocity observation instead of using actual joint velocities from the simulator.

\subsection{Symmetric Data Augmentation}\label{subsec:symmetric}

\begin{figure}[tb]
  \centering
  \includegraphics[scale=0.45]{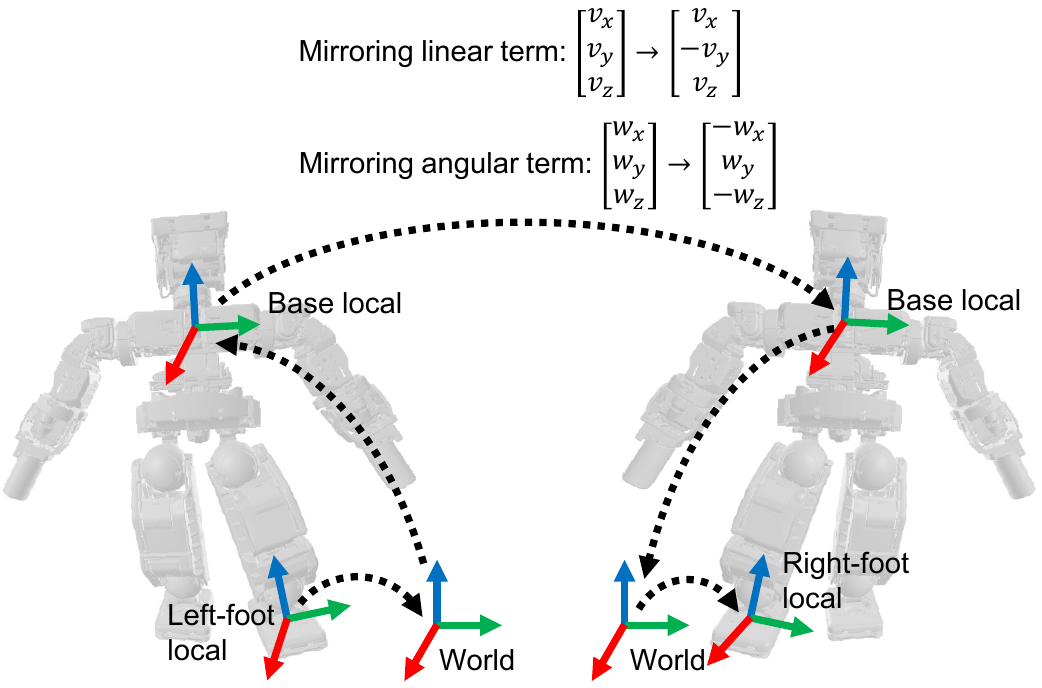}
  \caption{Coordinate frame transformations in mirroring left-foot IMU observations to right-foot IMU observations for symmetric data augmentation.}
  \label{fig:symmetric}
\end{figure}

We observe that naive RL can result in asymmetric and inefficient bipedal motions due to specific hardware design.
Specifically, in training the RL policy for EVAL-03 with our reward settings, naive RL tends to excessively avoid self-collisions between the left and right feet, which are very close even in the default joint position, as shown in Fig. \ref{fig:eval}.
The resultant motion can produce undesired yaw velocities due to the asymmetricity.
To enforce a symmetric policy with respect to the body center, we employ symmetric data augmentation \cite{mayank2024symmetry}. 

To generate symmetric observations, we mirror the given observations with respect to the body center.
To make the mirroring straightforward, most observation terms are expressed in the base local coordinate (e.g., velocity commands, projected gravity, base-mounted IMU measurements) or joint quantities.
However, we must perform coordinate frame transformations to mirror observations from the feet IMUs that are expressed in the local coordinate of each IMU.
These transformations are illustrated in Fig. \ref{fig:symmetric}.

\subsection{Teacher-Student Training with Fine-tuning}\label{subsec:teacherStudent}

\begin{figure}[t]
  \centering
  \includegraphics[width=\columnwidth]{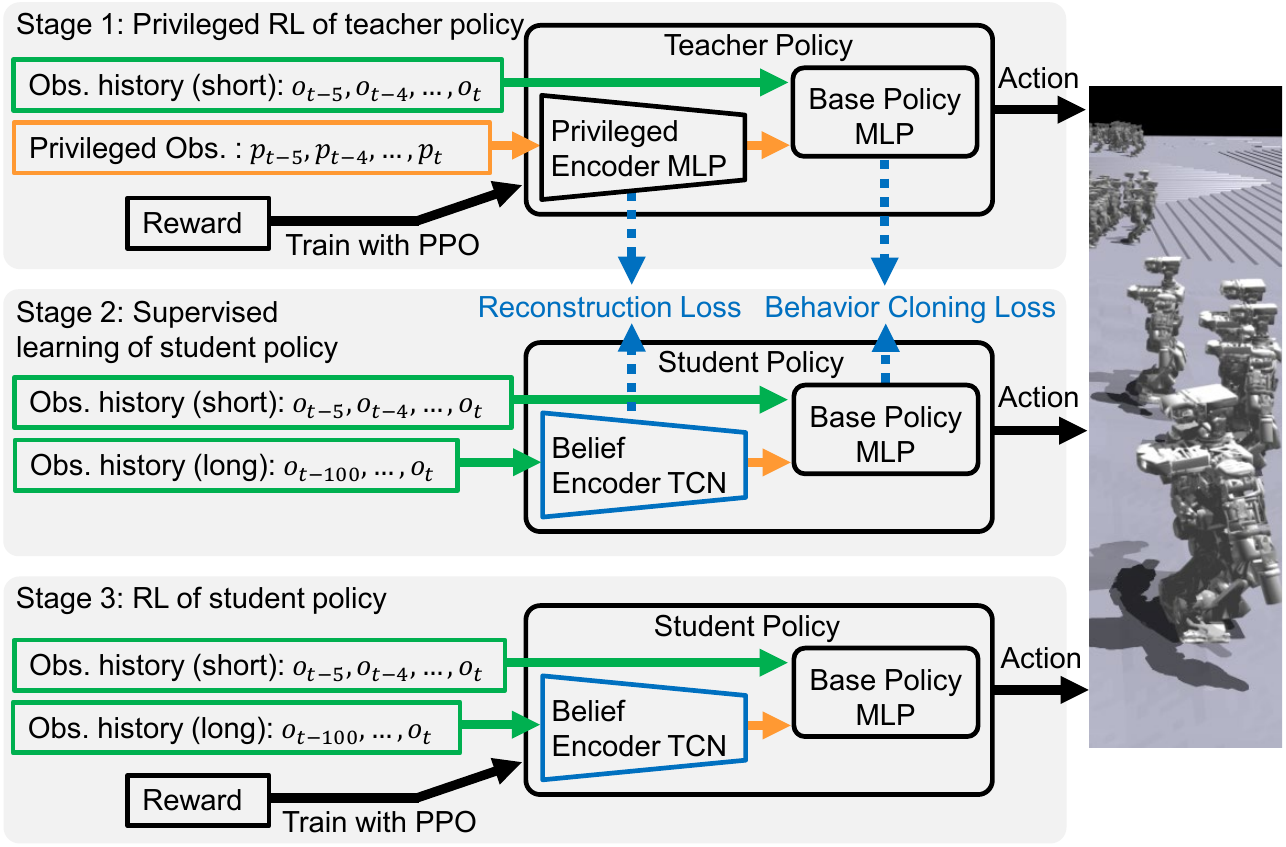}
  \caption{Teacher-student training with fine-tuning}
  \label{fig:rl}
\end{figure}

We adopt the teacher-student training framework for blind locomotion over rough terrains \cite{lee2020learning}, as illustrated in Fig. \ref{fig:rl}.
Initially, we train the teacher policy using a short observation history \((o_{t-4}, \ldots, o_t)\) and a short privileged observation history \((p_{t-4}, \ldots, p_t)\) using the Proximal Policy Optimization (PPO) algorithm \cite{schulman2017proximal}, with modifications for symmetric data augmentation \cite{mayank2024symmetry}.
The privileged observation terms are included in Table \ref{table:observations}.

Subsequently, we train the student policy through supervised learning to imitate the teacher's actions and reconstruct privileged information.
The student policy employs a temporal convolutional network (TCN) \cite{lea2017temporal} as a belief encoder to estimate the privileged information from the long observation history \((o_{t-99}, \ldots, o_{t})\).
During the supervised learning phase, we also leverage symmetric data augmentation: we collect data from the simulation using the student policy in an on-policy fashion, similar to DAgger, and augment its symmetric counterpart to the batch.

After the supervised learning of the student policy, we further fine-tune the student policy via RL using PPO.
In this phase, we employ an asymmetric actor-critic approach, providing privileged observations to the critic while withholding them from the actor.
In contrast to \cite{kumar2022adapting}, we train the entire student policy, as this approach enhances performance in our problem settings compared to fine-tuning only the base policy MLP.

\subsection{Random Network Distillation}\label{subsec:rnd}
During the RL training of the teacher policy, we utilize random network distillation (RND) \cite{burda2019exploration} to enhance exploration. 
We observe that, without RND, the teacher policy tends to exhibit minimal swing-foot clearance to overly avoid risks of falling, even we have the reward function to promote swing-foot clearance as listed in Table \ref{table:reward}. 
Following \cite{schwarke2023curiosity}, instead of using the full observations \(o_t\) in RND exploration, we define the so-called curiosity state \( s \) independent of the observations:
\begin{equation}\label{eq:curiosityState}
  s:= 
  \begin{bmatrix}
    r_{\rm left} \\ 
    r_{\rm right} \\ 
    {\rm HeightScan} \\ 
  \end{bmatrix},
\end{equation}
where \( r_{\rm left}, r_{\rm right} \in \mathbb{R}^3 \) denote the positions of the left and right feet expressed in the base local coordinate frame, respectively. 
The curiosity state defined in (\ref{eq:curiosityState}) aims to encourage exploration of various foot positions for each given terrain observation.
We choose MLPs with hidden sizes of (64, 64, 16) for the target network and (64, 32, 16) for the predictor network, respectively.
During training of a teacher policy, we add the intrinsic reward \cite{burda2019exploration} whose weight was set to 2.0 while updates the predictor network to reduce the difference between the outputs of the predictor and target networks.

\section{EXPERIMENTAL SETUP}\label{sec:experimental_setup}

To evaluate the effectiveness of the proposed method, we have compared the following three policies in the hardware experiments:
\begin{enumerate}
  \item Policy observing linear accelerations and angular velocities of base-mounted IMU and foot-mounted IMUs (\textbf{w/ Feet IMUs}) 
  \item Policy observing linear accelerations and angular velocities of base-mounted IMU (\textbf{w/o Feet IMUs 1})
  \item Policy observing angular velocities of base-mounted IMU (\textbf{w/o Feet IMUs 2})
\end{enumerate}
The first method represents our proposed approach, while the latter two represent existing methods. Through hardware experiments, we investigate how the additional feet IMU observations can mitigate sim-to-real gaps and enhance stability on real hardware.

\subsection{Hardware Details}
We use the gear-driven, miniature-sized humanoid robot EVAL-03, which is depicted in Fig. \ref{fig:eval}, throughout the experiments.
It stands approximately 240 mm tall from ground to the head link when standing at the default joint posture.
The total weight is around 1.73 kg.
The robot has 27 degrees of freedom (DOFs) in total: 6 DOFs in each leg, 3 DOFs in the torso, 4 DOFs in each arm, and 3 DOFs in the head.
However, in this paper, we treat the joints in the upper body as fixed joints for simplicity, reducing the total active DOFs to 12.

Consistent with its compact size, the motors are also small. Consequently, each actuator employs a high-gear ratio to compensate for the low-power motors while lacking a torque sensor.
To facilitate smooth sim-to-real transfer under these specifications, we employ high-gain PD control with the low pass filter (\ref{eq:LPF}) as introduced in \cite{haarnoja2024learning}.

The control architecture operates at multiple frequencies.
The policy runs at 100 Hz, while the low-level PD controller and orientation filter \cite{madgwick2010efficient} operate at 1000 Hz.
The orientation filter estimates the base rotation, expressed as a quaternion, from the base IMU observations, which is then converted to the projected gravity.

\subsection{Training Details}
For each policy, we trained multiple seeds and selected the best one for comparison through a three-stage process.
First, we trained six teacher policies with different seeds and selected the best two policies based on motion quality and reward performance.
Then, we trained four student policies with different seeds for each of the two selected teacher policies.
Finally, we selected the best one from the eight student policies based on sim-to-real transfer performance on the real hardware, rather than simulation reward values.

\subsubsection{Training Teacher Policy}
For teacher policy training, we collected trajectories using 4096 parallelized environments.
We implemented PPO with modifications for symmetric data augmentation \cite{mayank2024symmetry}.
The batch size was 196608 with symmetric data augmentation, and the number of minibatches was 6.
We utilized the adaptive learning rate as described in \cite{rudin2022learning}.
The teacher policy was trained for 20000 learning iterations.

\subsubsection{Training Student Policy}
For student policy training, we collected trajectories using 2048 parallelized environments.
The batch size was 98304 with symmetric data augmentation, and the number of minibatches was 6.
We employed a fixed learning rate of $5.0 \times 10^{-4}$.
The student policy was trained for 15000 learning iterations.

\subsubsection{Finetuning Student Policy}
For student policy finetuning, we collected trajectories using 2048 parallelized environments.
The batch size was 98304 with symmetric data augmentation, and the number of minibatches was 6.
We utilized the adaptive learning rate as described in \cite{rudin2022learning}.
The finetuning process continued for 25000 learning iterations.

\section{EXPERIMENTAL RESULTS}\label{sec:experimental_results}
\subsection{Walking on Floor}

\begin{table}[t]
\centering
\caption{Average walking speed for a given input velocity command in wallking on floor}
\begin{tabular}{lccc}
\toprule
\multirow{2}{*}{Method} & $v_{x, {\rm cmd}} = 0.05$ & $w_{z, {\rm cmd}} = 0.5$ & $w_{z, {\rm cmd}} = 1.0$ \\
                        & (forward)                 & (turn)                   & (fast turn)              \\
\midrule
\textbf{w/ Feet IMUs} & $v_{x} = 0.043$ & $w_{z} = 0.59$ & $w_{z} = 1.23$ \\
\midrule
w/o Feet IMUs 1       & $v_{x} = 0.033$ & $w_{z} = 0.71$ & -- \\
\midrule
w/o Feet IMUs 2       & $v_{x} = 0.03$  & $w_{z} = 0.66$ & --  \\
\bottomrule
\hline
\label{table:floor}
\end{tabular}
\end{table}

\begin{figure}[t]
  \begin{subfigure}[b]{\columnwidth}
    \centering
    \includegraphics[width=0.9\columnwidth]{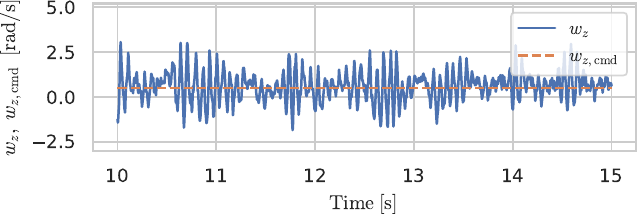}
    \caption{\textbf{w/ Feet IMUs}.}
  \end{subfigure}

  \vspace{1.5mm}
  \begin{subfigure}[b]{\columnwidth}
    \centering
    \includegraphics[width=0.9\columnwidth]{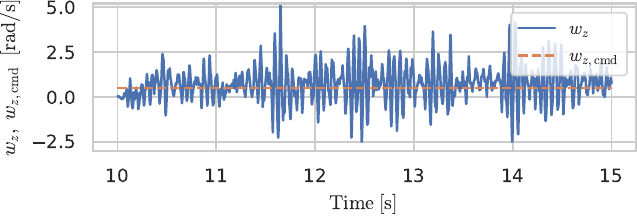}
    \caption{w/o Feet IMUs 1.}
  \end{subfigure}

  \vspace{1.5mm}
  \begin{subfigure}[b]{\columnwidth}
    \centering
    \includegraphics[width=0.9\columnwidth]{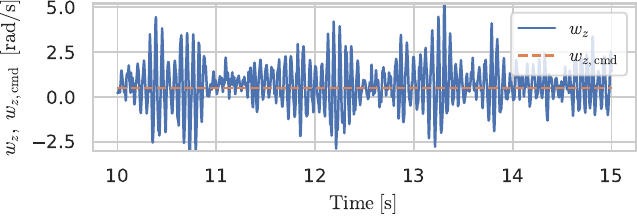}
    \caption{w/o Feet IMUs 2.}
  \end{subfigure}

  \caption{Plots of yaw velocities $w_{z}$ in tracking the turn command $w_{z, {\rm cmd}} = 0.5$ rad/s.}
  \label{fig:floor_turn_plots}
\end{figure}

\begin{figure}[t]
  \begin{subfigure}[b]{\columnwidth}
    \centering
    \includegraphics[width=0.925\columnwidth]{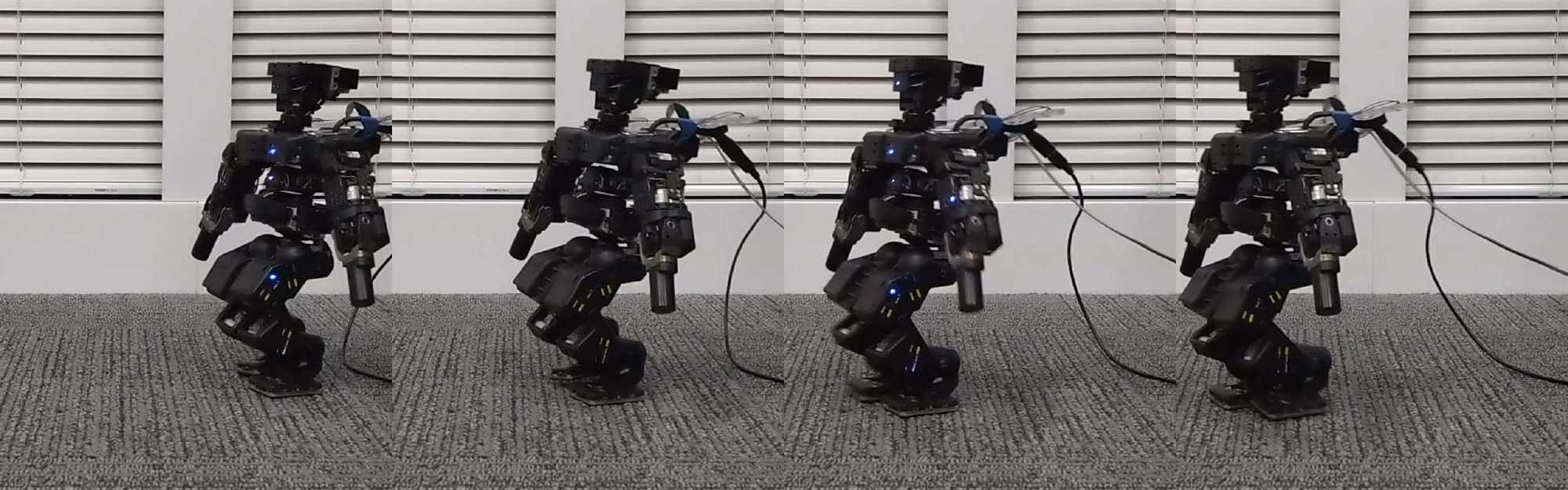}
    \caption{Forward command ($v_{x, {\rm cmd}} = 0.05$ m/s).}
  \end{subfigure}

  \vspace{1.5mm}
  \begin{subfigure}[b]{\columnwidth}
    \centering
    \includegraphics[width=0.925\columnwidth]{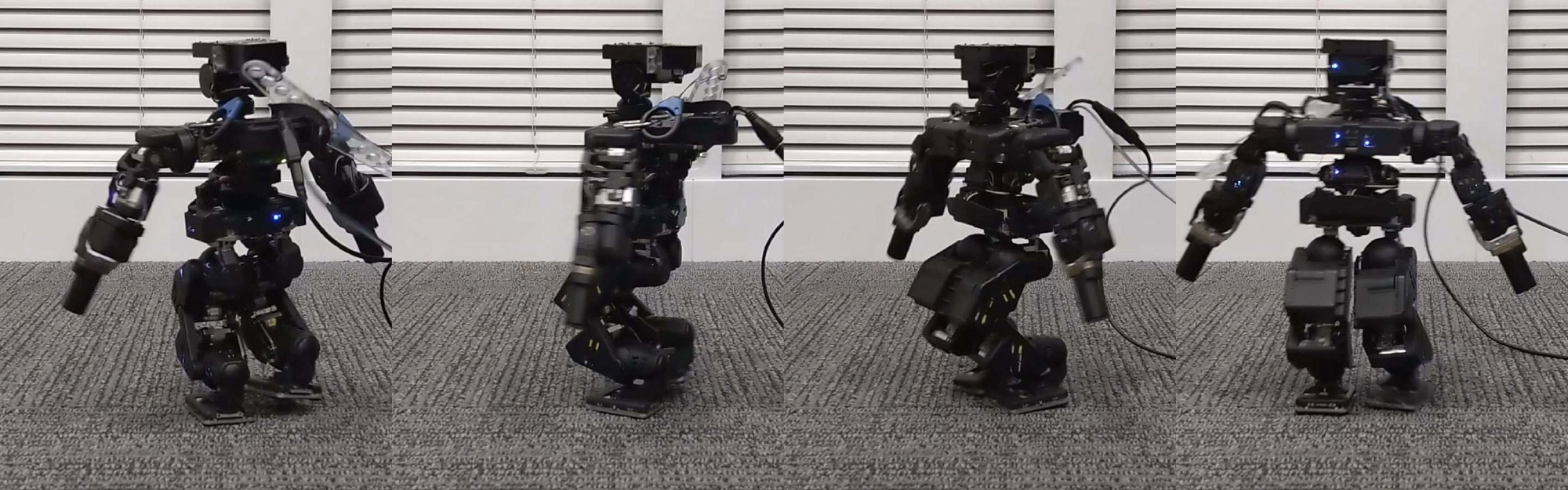}
    \caption{Fast turn command ($w_{z, {\rm cmd}} = 1.0$ rad/s).}
  \end{subfigure}

  \vspace{1.5mm}
  \begin{subfigure}[b]{\columnwidth}
    \centering
    \includegraphics[width=0.925\columnwidth]{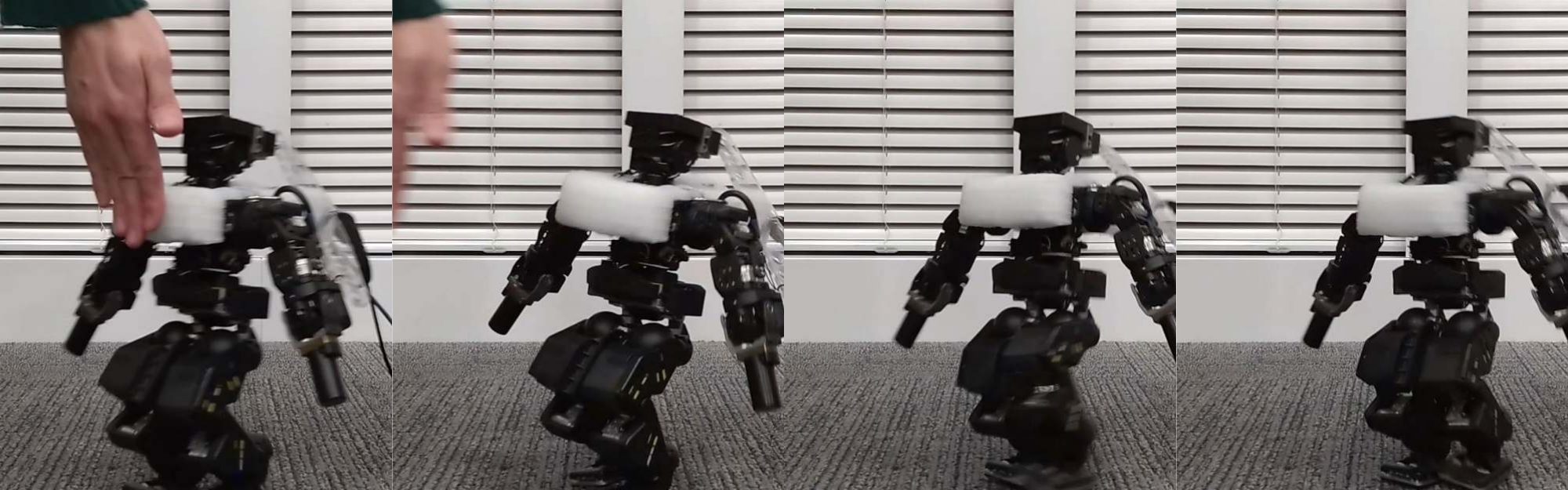}
    \caption{External push.}
  \end{subfigure}

  \caption{Snapshots of EVAL-03 walking on floor using the foot-mounted IMU observations.}
  \label{fig:walking_floor}
\end{figure}

First, we examined the performances of three policies on the floor to evaluate their velocity tracking capabilities.
We evaluated the performances with forward command ($v_{x, {\rm cmd}} = 0.05$ m/s), turn command ($w_{z, {\rm cmd}} = 0.5$ rad/s), and fast turn command ($w_{z, {\rm cmd}} = 1.0$ rad/s).

Table \ref{table:floor} shows the average walking speed of each policy for given velocity commands.
While the yaw velocity ($w_z$) was directly measured from the base-mounted IMU, the forward velocity of the robot ($v_x$) was estimated from videos because we could not measure it from equipped sensors.
As shown in Table \ref{table:floor}, the proposed method tracked the velocity commands better than the other methods in terms of average speed comparison. Notably, the proposed method could track the fast turn command while the other methods fell down by losing balance to track the fast yaw velocity command.
Fig. \ref{fig:floor_turn_plots} shows the plot of yaw velocity $w_z$ of each policy during tracking the turn command $w_{z, {\rm cmd}} = 0.5$, which also illustrates that the proposed method resulted in less deviation between $w_z$ from $w_{z, {\rm cmd}}$ than the other two policies.
Fig. \ref{fig:walking_floor} shows snapshots of EVAL-03 walking on the floor using the policy with foot-mounted IMU observations, including the reactive motion against an external push disturbance.
In the following experiments, we further compare such robustness among the three policies.

\subsection{Walking over a Variety of Terrains}

\begin{figure*}[htp]
    \centering
    \setlength{\tabcolsep}{2.0pt} % Adjust spacing between columns
    \renewcommand{\arraystretch}{0} % Adjust vertical spacing
    \begin{tabular}{ccccccc}
        \includegraphics[width=0.13\textwidth]{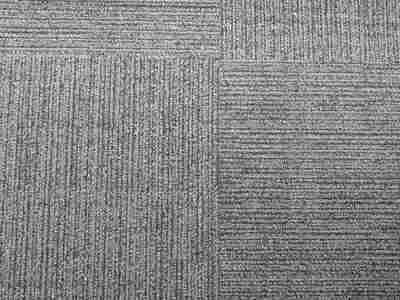} &
        \includegraphics[width=0.13\textwidth]{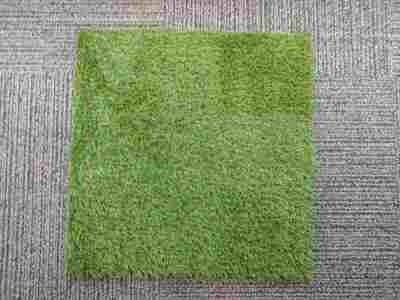} &
        \includegraphics[width=0.13\textwidth]{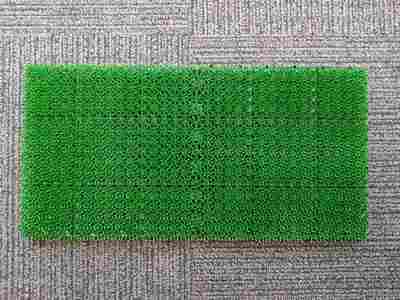} &
        \includegraphics[width=0.13\textwidth]{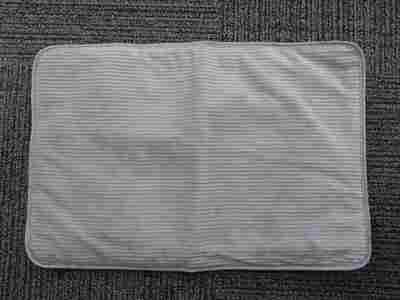} &
        \includegraphics[width=0.13\textwidth]{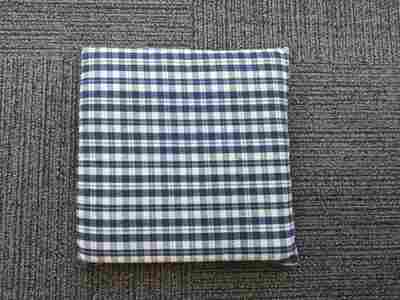} &
        \includegraphics[width=0.13\textwidth]{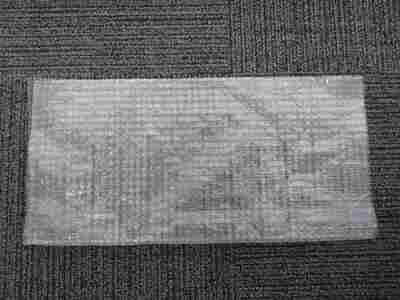} &
        \includegraphics[width=0.13\textwidth]{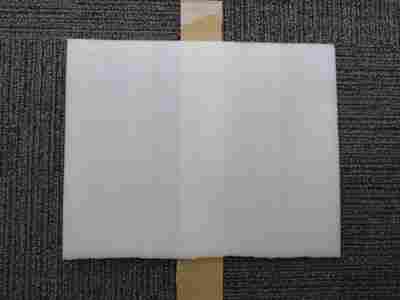} \\

        \vspace{1.0pt}
        \includegraphics[width=0.13\textwidth]{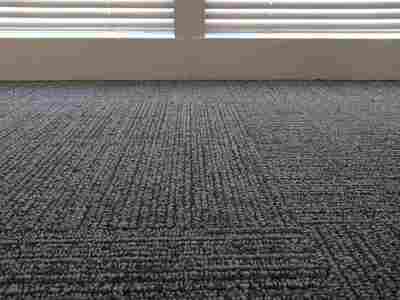} &
        \includegraphics[width=0.13\textwidth]{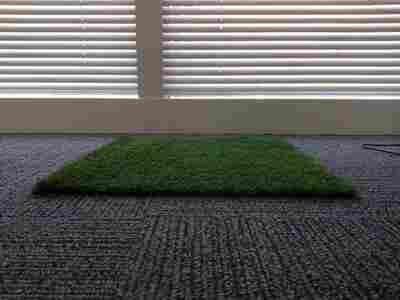} &
        \includegraphics[width=0.13\textwidth]{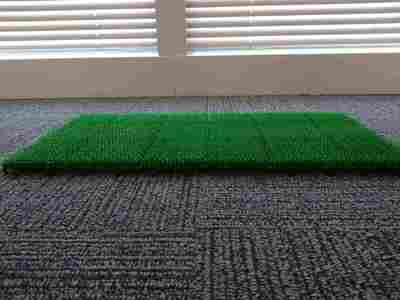} &
        \includegraphics[width=0.13\textwidth]{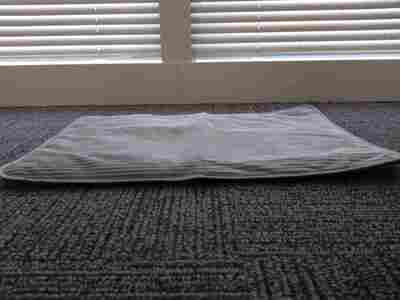} &
        \includegraphics[width=0.13\textwidth]{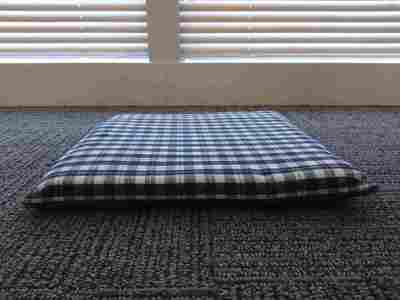} &
        \includegraphics[width=0.13\textwidth]{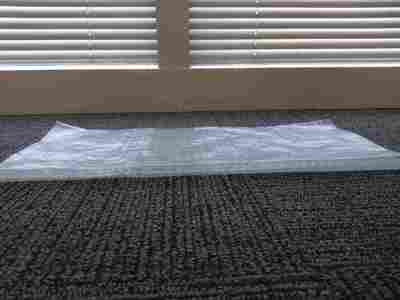} &
        \includegraphics[width=0.13\textwidth]{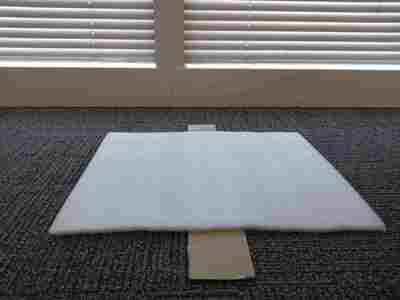}
    \end{tabular}
    \caption{Photos of terrains used in the hardware experiments: floort, turf (thin), turf (thick), cushion (pet), cushion (human), babble wrap, and uneven urethane sheet.}
    \label{fig:terrains}
\end{figure*}

Second, we examined the performance of the policies in walking over various terrains: floor, turf (thin), turf (thick), cushion (pet), cushion (human), bubble wrap, and uneven urethane sheet.
These terrains are depicted in Fig. \ref{fig:terrains}. We commanded forward walking with $v_{x, {\rm cmd}} = 0.05$ m/s, $v_{y, {\rm cmd}} = 0$ mm/s, and $w_{z, {\rm cmd}} = 0$ rad/s.
During the experiments, we measured two metrics: the success rate in traversing the terrain and the walking speed relative to the policy's performance on the floor.
Note that the walking speeds were only evaluated from successful cases.

\begin{table*}[t]
\centering
\caption{Success rates and speed rates in walking over a variety of terrains. The speed rates are computed by dividing the average moving speed over the terrain by the walking speed over the floor for each policy.}
\begin{tabular}{lcccccccc}
\toprule
Method & & Floor & Turf (thin) & Turf (thick) & Cushion (pet) & Cushion (human) & Babble wrap & Uneven urethane sheet \\
\midrule
\multirow{2}{*}{\textbf{w/ Feet IMUs}} & Success rate & 1.0 & 1.0  & 1.0  & \textbf{1.0}  & 0.0 & 1.0  & \textbf{0.8} \\
                                       & Speed rate   & 1.0 & 0.66 & 0.79 & \textbf{0.71} & --  & 1.13 & 0.77 \\
\midrule
\multirow{2}{*}{w/o Feet IMUs 1}       & Success rate & 1.0 & 1.0  & 1.0  & 0.4           & 0.0 & 1.0  & 0.0 \\
                                       & Speed rate   & 1.0 & 0.7  & 0.75 & 0.5           & --  & 1.15 & -- \\
\midrule
\multirow{2}{*}{w/o Feet IMUs 2}       & Success rate & 1.0 & 1.0  & 1.0  & 0.6           & 0.0 & 1.0  & 0.0 \\
                                       & Speed rate   & 1.0 & 0.64 & 0.71 & 0.25          & --  & 0.78 & -- \\
\bottomrule
\hline
\label{table:terrains}
\end{tabular}
\end{table*}

\begin{figure*}[t]
  \centering
  \vspace{1.5mm}
  \begin{subfigure}[t]{2\columnwidth}
    \centering
    \includegraphics[width=0.95\columnwidth]{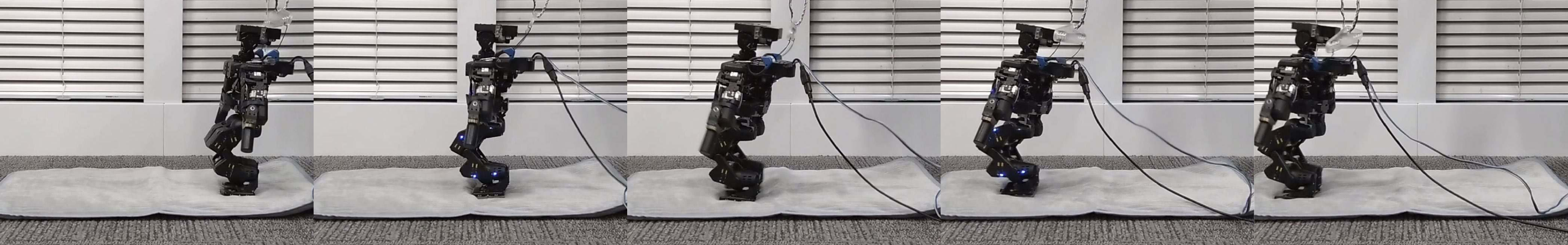}
    \caption{Cushion (pet).}
  \end{subfigure}

  \vspace{1.5mm}
  \begin{subfigure}[t]{2\columnwidth}
    \centering
    \includegraphics[width=0.95\columnwidth]{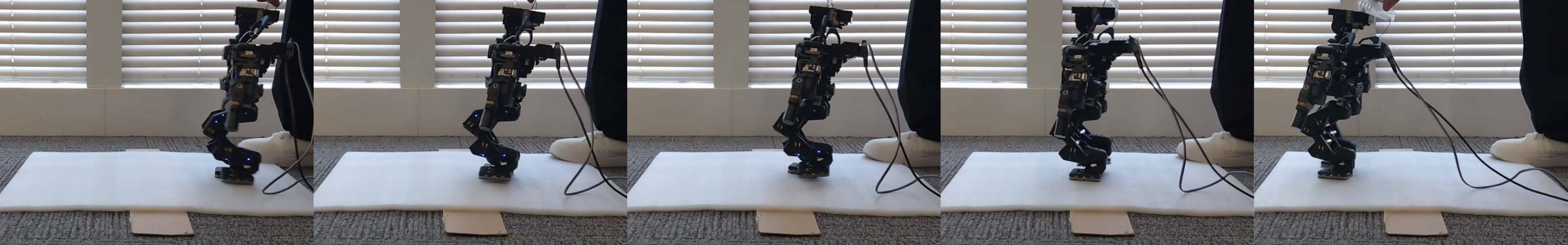}
    \caption{Uneven urethane sheet.}
  \end{subfigure}
  \caption{Snapshots of EVAL-03 walking over challenging terrains using the foot-mounted IMU observations.}
  \label{fig:terrain_walking}
\end{figure*}

Table \ref{table:terrains} shows the success rates and walking speed rates of the three policies, while Fig. \ref{fig:terrain_walking} shows snapshots of EVAL-03 walking over challenging terrains using the proposed method (w/ Feet IMUs).
As shown in Table \ref{table:terrains}, the feet IMU observations enhanced stability on uneven terrains.
Notably, for cushion (pet) and uneven urethane sheet, the proposed method achieved significantly higher success rates while the other policies failed to maintain balance and fell.
Additionally, the proposed method maintained consistent walking speeds even in the challenging cushion (pet) case comparable to floor walking, while other policies struggled and became stuck on soft terrains.

\subsection{Descending Steps}
Third, we evaluated the policies' performance in descending steps of varying heights (10 mm, 20 mm, and 25 mm).
We commanded forward walking with $v_{x, {\rm cmd}} = 0.05$ m/s, $v_{y, {\rm cmd}} = 0$ m/s, and $w_{z, {\rm cmd}} = 0$ rad/s.
We measured the success rate, defined as the percentage of successful step descents without falling.

\begin{table}[t]
\centering
\caption{Success rates in descending steps}
\begin{tabular}{lccc}
\toprule
Method                & Small & Medium & Large \\ 
\midrule
\textbf{w/ Feet IMUs} & 1.0         & 1.0          & 0.6  \\
\midrule
w/o Feet IMUs 1       & 0.0         & 0.2          & 0.0 \\
\midrule
w/o Feet IMUs 2       & 0.2         & 0.0          & 0.0  \\
\bottomrule
\hline
\label{table:steps}
\end{tabular}
\end{table}

\begin{figure}[t]
  \begin{subfigure}[b]{\columnwidth}
    \centering
    \includegraphics[width=0.925\columnwidth]{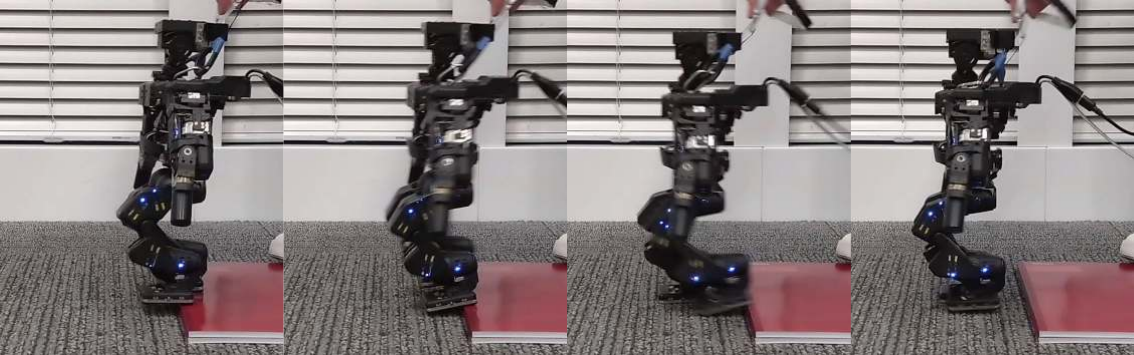}
    \caption{Small step (10 mm).}
  \end{subfigure}

  \vspace{1.5mm}
  \begin{subfigure}[b]{\columnwidth}
    \centering
    \includegraphics[width=0.925\columnwidth]{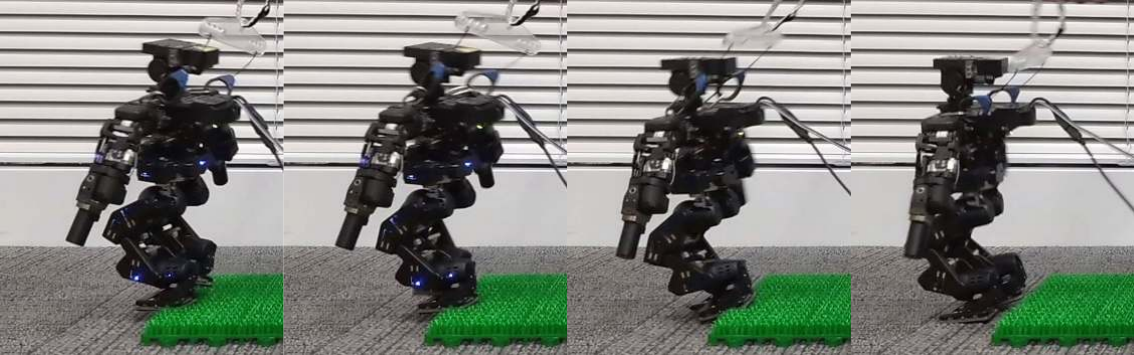}
    \caption{Medium step (20 mm).}
  \end{subfigure}

  \vspace{1.5mm}
  \begin{subfigure}[b]{\columnwidth}
    \centering
    \includegraphics[width=0.925\columnwidth]{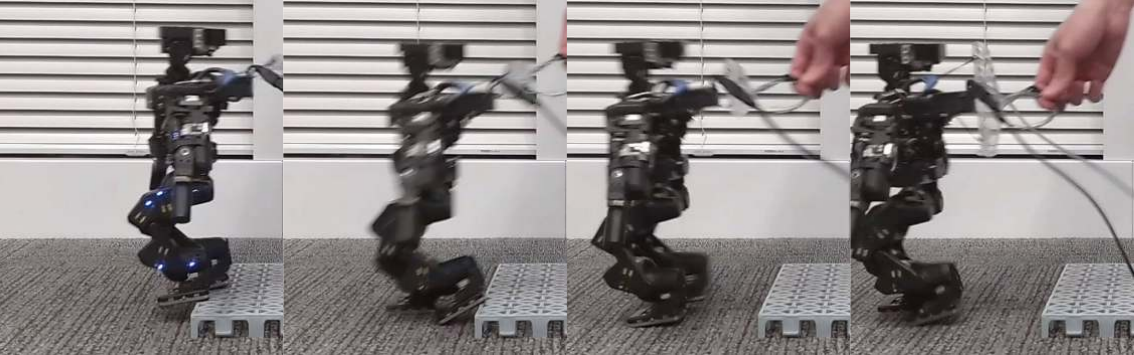}
    \caption{Large step (25 mm).}
  \end{subfigure}
  \caption{Snapshots of EVAL-03 stepping down from small (10 mm), medium (20 mm), and large (25 mm) steps using the foot-mounted IMU observations.}
  \label{fig:steps}
\end{figure}

Table \ref{table:steps} presents the success rates for step descent, while Fig. \ref{fig:steps} shows snapshots of EVAL-03 descending various steps using the proposed method (w/ Feet IMUs).
As shown in Table \ref{table:steps}, the proposed method successfully navigated steps where other methods consistently failed, demonstrating its enhanced robustness in sudden terrain transitions.

\subsection{Walking with Payloads}
Lastly, we evaluated the policies' performance while carrying unexpected payloads (0.33 kg and 0.55 kg).
We commanded forward walking with $v_{x, {\rm cmd}} = 0.05$ m/s, $v_{y, {\rm cmd}} = 0$ m/s, and $w_{z, {\rm cmd}} = 0$ rad/s and estimated the walking speed from the videos.

\begin{table}[t]
\centering
\caption{Forward walking speed $v_{x}$ [m/s] with unexpected payloads}
\begin{tabular}{lcc}
\toprule
Method                & 0.33 kg & 0.55 kg \\ 
\midrule
\textbf{w/ Feet IMUs} & 0.04    & 0.017 \\
\midrule
w/o Feet IMUs 1       & 0.008   & 0.012 \\
\midrule
w/o Feet IMUs 2       & 0.024   & 0.014 \\
\bottomrule
\hline
\end{tabular}
\label{table:payload}
\end{table}

\begin{figure}[t]
  \centering
  \includegraphics[width=0.925\columnwidth]{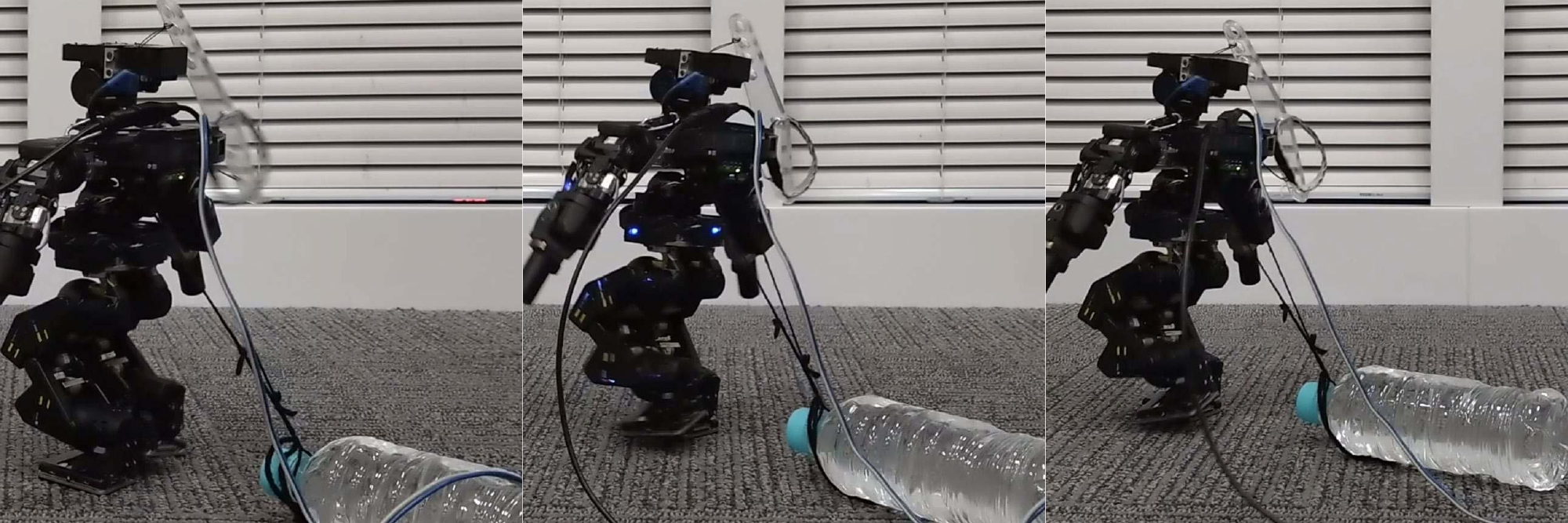}
  \caption{Snapshots of EVAL-03 walking with unexpected 0.55 kg payload using the foot-mounted IMU observations.}
  \label{fig:payload}
\end{figure}

Table \ref{table:payload} presents the walking speed with unexpected payloads, while Fig. \ref{fig:payload} shows snapshots of EVAL-03 walking with payloads using the proposed method (w/ Feet IMUs).
As shown in Table \ref{table:payload}, with a relatively light 0.33 kg payload (19 \% of the total mass), the proposed method maintained consistent walking speeds comparable to its unloaded performance, while other methods exhibited significant speed degradation when carrying payloads.
However, with a relatively heavy 0.55 kg payload (32 \% of the total mass), all three methods resulted in slow walking speeds.

\subsection{Discussion and Limitation}
The experimental results demonstrate that the proposed method exhibited rapid stabilization capabilities over challenging terrains, including non-rigid surfaces (cushion (pet) and uneven urethane sheet in Table \ref{table:terrains}) and sudden environmental transitions (step descent in Table \ref{table:steps}).
We hypothesize that foot-mounted IMUs enable direct and rapid measurement of feet states, which helps the policy cope with balance challenges arising from contacts with various environments.

However, several limitations remain.
The proposed method failed to maintain balance in more challenging scenarios (e.g., human-sized cushion in Fig. \ref{fig:terrains}).
Furthermore, the proposed method was unable to climb even modest obstacles, such as a 5 mm step, as well as the other two policies.
This limitation suggests that terrain feature estimation solely through foot-mounted IMUs may be insufficient within our current learning framework despite utilizing a long observation history of up to 100 time steps (1.0 s).
Alternative approaches, such as utilizing joint position tracking errors with low-gain PD control \cite{lee2020learning}, still remain promising for addressing these challenges.

\section{CONCLUSIONS}\label{sec:conclusions}
This paper presented a novel approach to learning bipedal locomotion on gear-driven humanoid robots using foot-mounted IMUs.
Rather than pursuing complex actuator modeling or system identification, we introduced linear acceleration and angular velocity measurements from foot-mounted IMUs as well as the base-mounted IMU within the blind locomotion learning framework.
We also introduced symmetric data augmentation and random network distillation to enhance bipedal locomotion learning with the proposed framework.
Through hardware experiments on EVAL-03 with a variety of settings, we showed that the proposed method improved stability on non-rigid surfaces and during sudden environmental transitions, such as step descents.
However, limitations remain, particularly in upward step navigation, which suggests directions for future research to introduce compliant joint control with lower PD gains.

\bibliographystyle{IEEEtran}
\bibliography{IEEEabrv, root}

\end{document}